# Real-Time Trajectory Planning for AGV in the Presence of Moving Obstacles: A First-Search-Then-Optimization Approach *

Bai Li, Youmin Zhang, Yakun Ouyang, Yi Liu, Xiang Zhong, Hangjie Cen, Qi Kong

*Abstract*—**This paper focuses on automatic guided vehicle (AGV) trajectory planning in the presence of moving obstacles with known but complicated trajectories. In order to achieve good solution precision, optimality and unification, the concerned task should be formulated as an optimal control problem, and then discretized into a nonlinear programming (NLP) problem, which is numerically optimized thereafter. Without a near-feasible or near-optimal initial guess, the NLP-solving process is usually slow. With the purpose of accelerating the NLP solution, a search-based rough planning stage is added to generate appropriate initial guesses. Concretely, a continuous state space is formulated, which consists of Cartesian product of 2D configuration space and a time dimension. The rough trajectory is generated by a graph-search based planner, namely the A* algorithm. Herein, the nodes in the graph are constructed by discretizing the aforementioned continuous spatio-temporal space. Through this first-search-then-optimization framework, optimal solutions to unified trajectory planning problems can be obtained fast. Simulations have been conducted to verify the real-time performance of our proposal.**

## I. INTRODUCTION

### A. Background and Motivations

An automated guided vehicle or automatic guided vehicle (AGV) is a mobile robot that moves on the floor under the guidance of navigation information [1]. Ever since their emergence in the 1950s, AGVs have become a promising means of organizing intralogistics in various applications such as mining, surveillance, forklift automation, satellite clustering, and warehousing [2,3]. Trajectory planning, as a module in an AGV system, is responsible for generating the AGV's running motion at every subtle moment. Trajectory planning is usually the direct reflection of an AGV's intelligence level [4]. This paper focuses on the AGV trajectory planning scheme.

Commonly the workspace of an AGV also contains other movable objects, which may be cooperative or non-cooperative. If the other objects are cooperative, a centralized system would plan collision-free trajectory for each object, including the AGV [5]. Conversely, if the objects are non-cooperative, the AGV should avoid collisions with the other moving objects. Compared with cooperative trajectory planning, non-cooperative trajectory planning is more difficult because the degree of freedom is smaller [6,7]. This work deals with the AGV trajectory planning problem with non-cooperative moving objects, which is often referred to as trajectory planning in the dynamic environment or with moving obstacles [8]. Even if the sensor range limits and interior/exterior uncertainties are ignored, the concerned problems are still difficult, i.e., being NP-hard [9].

### B. Related Works

The predominant AGV trajectory planners which can handle moving obstacles are classified as direct and indirect methods. A direct method generates a collision-free and kinematically feasible trajectory all at once. By contrast, an indirect method regards a trajectory as a combination of multiple decoupled phases, and makes plan for each phase in parallel or in sequence.

Typical indirect planners include the path-velocity decomposition (PVD) method, which first plan a path, and then attach a feasible velocity profile along the path [10–12]. Indirect methods usually run fast, because the difficulties in the original trajectory planning problem are dispersed into multiple phases. However, indirect methods are commonly incomplete because the original problem is transformed into an inequivalent form. For example, the PVD method would fail if there is no collision-free velocity along the planned path. Such a failure is typically known as deadlock in robotics.

Direct planners primarily consist of the search-based, sample-based, and optimal-control-based methods. Shih et al. [13] proposed a hierarchical framework, wherein a graph searcher plans a rough trajectory in the space-time domain, and precise optimization is done thereafter. Ziegler and Stiller [14] developed a deterministic sample-based planner called spatio-temporal state lattice, which is an extension of the conventional state lattice path planners. Hsu et al. [15] proposed a stochastic sample-based planner by extending the conventional probabilistic roadmap (PRM) framework. Qu et al. [16] formulated the concerned trajectory planning task as an optimal control problem (OCP), and then provided feasible solutions after converting the OCP into a chained-form system. Ref. [17] introduced additional degrees of freedom in the OCP formulated in [16], and then applied the sequential quadratic programming (SQP) algorithm to find near-optimal solutions. The search-based and sample-based planners usually generate solutions in the abstracted space-time domain, thus the solutions, although being rough, are derived fast. On the other hand, the optimal-control-based planners can provide precise solutions, but the computational burden is heavy. With the purpose of finding precise and optimal solutions fast, there has been a trend, as in [13], to use the rough trajectory obtained by the search-based or sample-based planner as the initial guess

* This work is supported by the Fundamental Research Funds for the Central Universities under Grant 531118010509.

Bai Li is with the College of Mechanical and Vehicle Engineering, Hunan University, Changsha, China (e-mail: libai@hnu.edu.cn).

Youmin Zhang is with the Department of Mechanical, Industrial and Aerospace Engineering, Concordia University, Montreal, Canada (e-mail: ymzhang@encs.concordia.ca).

Qi Kong is with the JDX R&D Center, JD.com American Technologies Corporation, Mountain View, CA (e-mail: spiritkong@gmail.com).

Yakun Ouyang, Hangjie Cen, and Xiang Zhong are with the College of Mechanical and Vehicle Engineering, Hunan University, Changsha, China.

for warmly starting the numerical optimization process in a precisely formulated OCP [18,19]. However, the pioneering work [13] did not consider the nonholonomic constraints related to the AGV kinematics, and restricted the obstacles' trajectories to be piecewise constant.

To summarize, there has not been a unified method that can deal with generic AGV trajectory planning cases with arbitrarily assigned trajectories in a real-time, optimal, and precise way.

### *C. Contribution*

In this work, the AGV trajectory planning scheme with arbitrarily specified moving obstacles is formulated as a unified OCP. The formulated OCP is discretized into a nonlinear programming (NLP) problem, and then numerically optimized thereafter. With the purpose of solving the NLP problem fast, a search-based rough planning stage is developed. The preliminary trajectory derived at that rough stage serves as the initial guess for warmly starting the NLP optimization process. Although being rough, the fast generated initial guess can largely reduce the NLP solution time. Through this first-rough-then-precise computation framework, real-time solutions to generic trajectory planning schemes are derived.

The remainder of this paper is organized as follows. Section II briefly states the concerned trajectory planning task. Rough and precise solution strategies are introduced in Sections III and IV, respectively. Section V provides simulation results and discussions. Finally, Section VI concludes the paper.

## II. Problem Statement

This section defines the generic AGV trajectory planning task as an optimal control problem, which is about minimizing a specified cost function, subject to the kinematic constraints, collision-avoidance constraints, and boundary conditions.

### *A. AGV Kinematics*

Suppose the concerned AGV has front steering wheels. Commonly, the vehicle kinematics can be described as the following nonholonomic constraints [6]:

$$\frac{\mathrm{d}}{\mathrm{d}t}\begin{bmatrix} x(t) \\ y(t) \\ v(t) \\ \phi(t) \\ \theta(t) \end{bmatrix} = \begin{bmatrix} v(t)\cdot\cos\theta(t) \\ v(t)\cdot\sin\theta(t) \\ a(t) \\ \omega(t) \\ v(t)\cdot\tan\phi(t)/\mathrm{L_W} \end{bmatrix},\ t\in[0,\mathrm{t_f}]. \tag{1a}$$

Herein, $t$ is the time index, $\mathrm{t_f}$ is the specified terminal moment, $(x, y)$ refers to the mid-point of rear wheel axis (see point $P$ in Fig. 1), $\theta$ refers to the orientation angle, $v$ refers to the velocity of point $P$, $a$ refers to the corresponding acceleration, $\phi$ refers to the steering angle of the front wheels, $\omega$ refers to the corresponding angular velocity, and $\mathrm{L_W}$ denotes the wheelbase length. Several boundaries are imposed to restrict the movement of the vehicle during the entire movement process $t\in[0,\mathrm{t_f}]$:

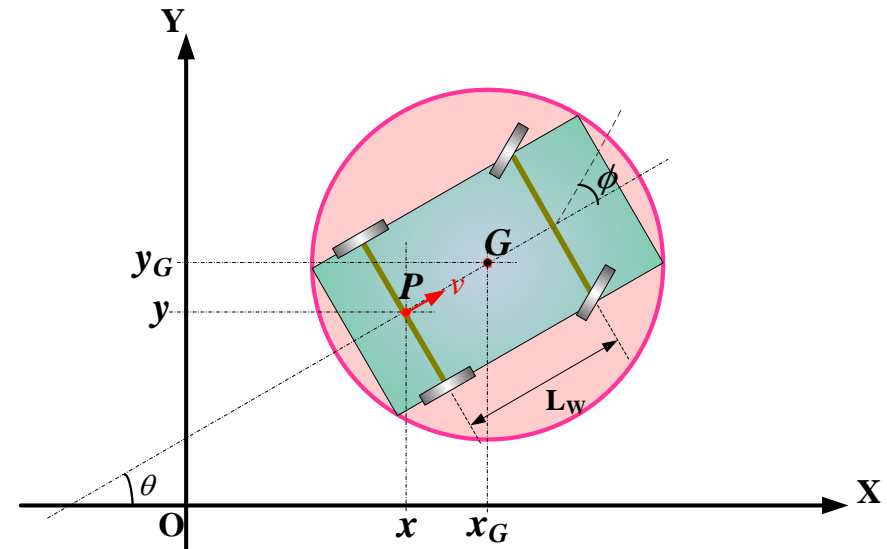

Fig. 1. Parametric notations related to vehicle shape and kinematics.

$$|a(t)| \le \mathrm{a_{max}}, \tag{1b}$$

$$|v(t)| \le \mathrm{v_{max}}, \tag{1c}$$

$$|\phi(t)| \le \Phi_{\mathrm{max}}, \tag{1d}$$

$$|\omega(t)| \le \Omega_{\mathrm{max}}, \tag{1e}$$

where $\mathrm{a_{max}}$, $\mathrm{v_{max}}$, $\Phi_{\mathrm{max}}$, and $\Omega_{\mathrm{max}}$ represent the upper bounds of the corresponding variables, respectively.

### *B. Collision Avoidance Constraints*

This work holds a fundamental assumption that the location of each moving/static obstacles in the environment in the future time interval $[0,\mathrm{t_f}]$ can be correctly predicted. Suppose that each obstacle is circular, the location of the *i*th obstacle's circle center $\left(\mathrm{obs_x}_i(t),\mathrm{obs_y}_i(t)\right)$ and radius $r_i(t)$ at any moment $t\in[0,\mathrm{t_f}]$ is known ( $i=1,\ldots,\mathrm{N_{obs}}$ , where $\mathrm{N_{obs}}$ denotes the number of obstacles).

Nominally the car-like ego-AGV is rectangular in the shape, but the corresponding collision-avoidance constraints are overly complicated. As a common practice, a circle is used to cover the rectangular body of the AGV [20]. Herein, the center of the circle is the geometric center of the AGV (see point $G=(x_G, y_G)$ in Fig. 1), and the circle radius $\mathrm{R_{ego}}$ can be easily determined according to the AGV's rectangular shape. The collision-avoidance constraints are formulated as

$$\begin{aligned} &\left(x_G(t)-\mathrm{obs_x}_i(t)\right)^2+\left(y_G(t)-\mathrm{obs_y}_i(t)\right)^2 \\ &\ge\left(\mathrm{R_{ego}}+r_i(t)\right)^2,\ i=1,\ldots,\mathrm{N_{obs}},\ t\in[0,\mathrm{t_f}]. \end{aligned} \tag{2}$$

### *C. Two-Point Boundary Conditions*

Boundary conditions consist of the configuration specifications at $t=0$ and $t=\mathrm{t_f}$. Concretely we have

$$\begin{aligned} &\left[x(0), y(0), \theta(0), v(0), \phi(0), a(0), \omega(0)\right]= \\ &\left[\mathrm{x_0}, \mathrm{y_0}, \theta_0, \mathrm{v_0}, \phi_0, \mathrm{a_0}, \omega_0\right], \end{aligned} \tag{3}$$

where $\mathrm{x_0}$, $\mathrm{y_0}$, $\theta_0$, $\mathrm{v_0}$, $\phi_0$, $\mathrm{a_0}$, and $\omega_0$ are parameters which determine the starting configurations. The terminal configurations can be specified in a same way, but commonly one may need to relax the terminal condition because it is not known *a priori* whether the ego-AGV is really able to reach a specified ending point at $t=\mathrm{t_f}$. In this work, we consider partly relax the terminal condition to the cost function. For example, we may only impose equalities on $\left[v(\mathrm{t_f}), \phi(\mathrm{t_f}), a(\mathrm{t_f}), \omega(\mathrm{t_f})\right]$:

$$\left[v(\mathrm{t_f}), \phi(\mathrm{t_f}), a(\mathrm{t_f}), \omega(\mathrm{t_f})\right] = \left[\mathrm{v_f}, \phi_\mathrm{f}, \mathrm{a_f}, \omega_\mathrm{f}\right]. \quad (4)$$

Next subsection introduces a way to handle the rest few states $\left[x(\mathrm{t_f}), y(\mathrm{t_f}), \theta(\mathrm{t_f})\right]$.

### D. Cost Function and Overall Problem Formulation

Generally speaking, the cost function $J$ consists of the terms that are expected to be minimized. Following the example in the preceding subsection, if we expect the terminal configurations $\left[x(\mathrm{t_f}), y(\mathrm{t_f}), \theta(\mathrm{t_f})\right]$ of the ego- AGV is close to a specified one $\left[\mathrm{x_f}, \mathrm{y_f}, \theta_\mathrm{f}\right]$, then we have

$$J = \left(x(\mathrm{t_f}) - \mathrm{x_f}\right)^2 + \left(y(\mathrm{t_f}) - \mathrm{y_f}\right)^2 + \left(\theta(\mathrm{t_f}) - \theta_\mathrm{f}\right)^2, \quad (5)$$

which leads the ego-AGV towards $\left[\mathrm{x_f}, \mathrm{y_f}, \theta_\mathrm{f}\right]$ at $t = \mathrm{t_f}$.

With all the aforementioned elements summarized, a standard AGV trajectory planning problem is defined as the following fixed-time optimal control problem.

$$\begin{aligned} &\min\ J, \\ &\text{s.t. Kinematic constraints (1);} \\ &\quad\ \text{Collision-avoidance constraints (2);} \\ &\quad\ \text{Initial conditions (3), and terminal conditions.} \end{aligned} \quad (6)$$

The unknown variables include $x(t)$, $y(t)$, $\theta(t)$, $v(t)$, $a(t)$, $\omega(t)$, and $\phi(t)$, $t \in [0, \mathrm{t_f}]$.

## III. Rough Trajectory Planning

A graph-search based trajectory planner, i.e. a modified A* algorithm, is introduced in this section for the generation of a rough trajectory. Compared with the conventional A* algorithm, our proposal makes changes in the following aspects: (i) time is regarded as the third dimension in addition to the 2D plane; (ii) state expansion in the dimension of time is strictly monotonous; (iii) the goal node is not a specified point in the *x*-*y*-*t* state space, but a manifold only with specified *t*. In the rest of this paper, we refer to our modified A* algorithm as 3D A* algorithm, the details of which are presented as follows.

As the preliminary step of the 3D A* algorithm, a continuous *x*-*y*-*t* state space is formulated, and then abstracted uniformly in each of the dimensions [21–23]. Through this, the continuous *x*-*y*-*t* state space is represented by a finite number of grids. Each grid in the abstracted space is called a *node*. The nodes occupied by the moving obstacles should be marked. The starting node A and the goal node B should be specified.

Like the conventional A* algorithm, 3D A* algorithm maintains two functions, i.e. $g(s)$ and $f(s)$, that map from a state $s$ to a scalar. Concretely, $g(s)$ measures the cost of the path from the starting node A to $s$, while $f(s) = g(s) + h(s)$ estimates the gross cost from A to the goal node B via $s$. In addition, 3D A* algorithm maintains a priority queue called OPEN, wherein the to-be-expanded nodes are recorded in a ascending sequence of $f(s)$. 3D A* algorithm begins from initializing OPEN with node A. Once a valid state $s$ in OPEN is expanded, it is removed from OPEN, and its valid successors are emplaced in OPEN. For each successor $s'$, if it has not been expanded before, it is added in OPEN with its parent set as $s$, and $f(s') = g(s) + c(s, s') + h(s')$, wherein $c(s, s')$ measures the cost between adjacent nodes $s$ and $s'$. If $s'$ is already in OPEN, it indicates that $s'$ has previously got its $f(s')$ value. In that case, we should try $f^*(s') = g(s) + c(s, s') + h(s')$: if $f^*(s') < f(s')$, the parent of $s'$ is updated as $s$; otherwise nothing needs to be done. Any successor $s'$ that used to be expanded but has been removed from OPEN should be ignored. This iterative process continues until OPEN is empty (which indicates a failure), or the terminal condition is satisfied (which indicates a success). Commonly, the terminal condition is that the goal node B gets expanded. But our 3D A* algorithm makes a relaxation, i.e., if any expanded node reaches B only in the time dimension, then the search process is terminated.

The pseudocode of 3D A* algorithm is given as follows.

**Algorithm 1.** 3D A* Algorithm.

**Input:** Abstracted *x*-*y*-*t* state space with occupied nodes marked, starting node $\mathrm{A} = (\mathrm{x_0}, \mathrm{y_0}, \mathrm{t_0})$, and goal node $\mathrm{B} = (\mathrm{x_f}, \mathrm{y_f}, \mathrm{t_f})$;

**Output:** Feasible path in the spatio-temporal space;

1. Set $\mathrm{OPEN} = \varnothing$, $\mathrm{CLOSED} = \varnothing$;
2. Set $g(\mathrm{A}) = 0$, calculate $f(\mathrm{A})$, and emplace node A into OPEN;
3. Find the node $\mathrm{S} = (\mathrm{x_S}, \mathrm{y_S}, \mathrm{t_S})$ with minimum $f(\mathrm{S})$ in OPEN;
4. **While** $\mathrm{t_S} \neq \mathrm{t_f}$, **do**
5. Remove S from OPEN to CLOSED;
6. **For** each successor $\mathrm{S'}$ of S, **do**
7. **If** $\mathrm{S'}$ is in the scope of the abstracted state space, and is not occupied by obstacles, and is not in CLOSED, **then**
8. **If** $\mathrm{S'}$ is not in OPEN, **then**
9. Add $\mathrm{S'}$ to OPEN;
10. Set S as the parent of $\mathrm{S'}$;
11. Calculate and record $f(\mathrm{S'}) = g(\mathrm{S}) + c(\mathrm{S}, \mathrm{S'}) + h(\mathrm{S'})$;
12. **Else**
13. Calculate $f^*(\mathrm{S'}) = g(\mathrm{S}) + c(\mathrm{S}, \mathrm{S'}) + h(\mathrm{S'})$;
14. **If** $f^*(\mathrm{S'}) < f(\mathrm{S'})$, **then**
15. Update $f(\mathrm{S'})$ as $f^*(\mathrm{S'})$;
16. Reset S as the parent of $\mathrm{S'}$;
17. **End if**
18. **End if**
19. **End if**
20. **End for**
21. **If** $\mathrm{OPEN} = \varnothing$, **then**
22. Return with failure reported;
23. **Else**
24. Find the node $\mathrm{S} = (\mathrm{x_S}, \mathrm{y_S}, \mathrm{t_S})$ with minimum $f(\mathrm{S})$ in OPEN;
25. **End if**
26. **End while**
27. Backtrack the nodes from the lasted expanded one to A, then inversely form a path in the *x*-*y*-*t* space;
28. Return with the derived path.

As presented in Line 4 of Algorithm 1, the obtained trajectory may not reach the specified goal node B. This relaxed terminal condition makes sense because we generally do not know whether the AGV can really reach B. In spite of this relaxation, the trajectory obtained by Algorithm 1 tries to reach node B due to the heuristics in $h(\cdot)$ .

## IV. PRECISE TRAJECTORY PLANNING

This section briefly introduces how to solve the formulated optimal control problem (i.e. (6)) numerically. First, (6) is discretized into an NLP problem using the first-order explicit Runge-Kutta method. Thereafter, the resulting NLP problem is optimized using the interior-point method (IPM) [5]. Particularly, the rough trajectory generated by the 3D A* method is used to warmly start the NLP optimization process. The output of the precise optimization stage is the final trajectory.

## V. SIMULATION RESULTS AND DISCUSSIONS

Simulations were performed in the AMPL/MATLAB platform and executed on an i5-7200U CPU with 8 GB RAM that runs at 2.50 $\times 2$ GHz. For the convenience of presentation, the proposed first-search-then-optimization algorithm is abbreviated as FSTO. The rough planning stage is referred to as Stage 1, and the precise optimization stage is regarded as Stage 2. Basic parametric settings are listed in Table I.

TABLE I. PARAMETRIC SETTINGS REGARDING MODEL AND APPROACH.

| Parameter | Description | Setting |
|---|---|---|
| $L_F$ | Front hang of vehicle | 0.707 m |
| $L_W$ | Wheelbase of vehicle | 1.414 m |
| $L_R$ | Rear hang of vehicle | 0.707 m |
| $L_B$ | Width of vehicle | 1.414 m |
| $R_{ego}$ | Starting configuration | 2.0 m |
| $a_{max}$ | Bound of $\lvert a(t)\rvert$ | 1.0 m/s$^2$ |
| $v_{max}$ | Bound of $\lvert v(t)\rvert$ | 1.0 m/s |
| $\Phi_{max}$ | Bound of $\lvert \phi(t)\rvert$ | 0.35 rad |
| $\Omega_{max}$ | Bound of $\lvert \omega(t)\rvert$ | 0.3 rad/s |
| NE | Discretization precision in Runge-Kutta method | 80 |

### A. *On the Efficiency of FSTO*

This subsection illustrates the procedures in FSTO via one simulation case, wherein the ego-AGV and three moving obstacles run in a 40 m $\times$ 10 m map. Let us set $t_f = 40.0$ sec . The trajectory of each moving obstacle during $t \in [0, t_f]$ is set as a Bézier curve with 4 random knots (Fig. 2). The radius of each circular obstacle is set as a random value from 0.5 m to 1.0 m. The starting configuration $[x_0, y_0, \theta_0, v_0, \phi_0, a_0, \omega_0]$ is [0,5,0,0,0,0,0], while the ego-AGV is expected to reach the goal $x(t_f) = 40$, $y(t_f) = 5$ with $\theta(t_f) = v(t_f) = \phi(t_f) = 0$ . Particularly, the AGV is not allowed to drive backward, thus only four expansion directions $(\Delta x, \Delta y, \Delta t)$ are considered: (1,-1,1), (1,1,1), (1,0,1), and (0,0,1).

The continuous *x*-*y*-*t* state space is presented in Fig. 3(a), which is abstracted into $40 \times 10 \times 40$ nodes (see Fig. 3(b)). 3D A* algorithm is applied to search for a rough trajectory in the abstracted grid space, and the derived rough trajectory (Fig. 4(a)) is utilized to initialize the NLP solving process at the precise optimization stage. The optimized trajectory is shown in Fig. 4(b). Compared with the rough trajectory, the precisely optimized trajectory is smoother, and both trajectories are homotopic with each other (Fig. 5). In addition, two more complicated cases are depicted in Figs. 6 and 7.

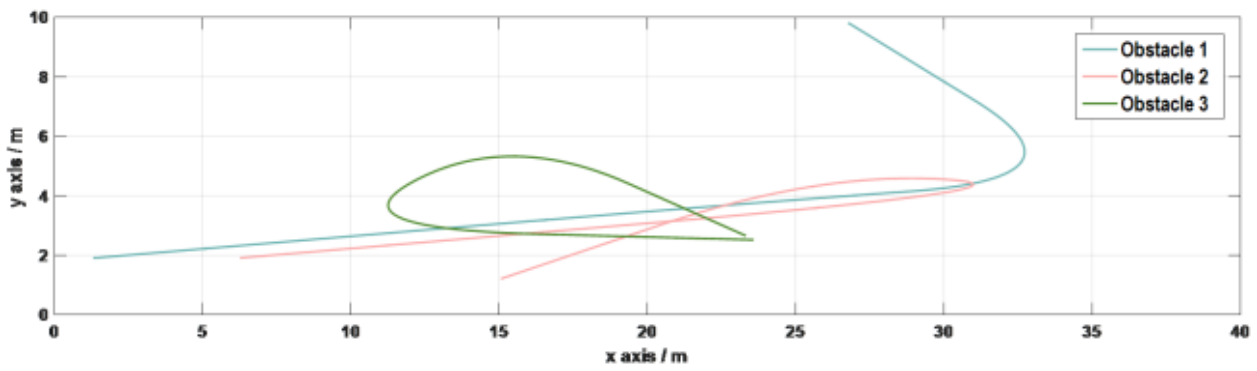

Fig. 2. Trajectories of three moving obstacles during $t \in [0, t_f]$ .

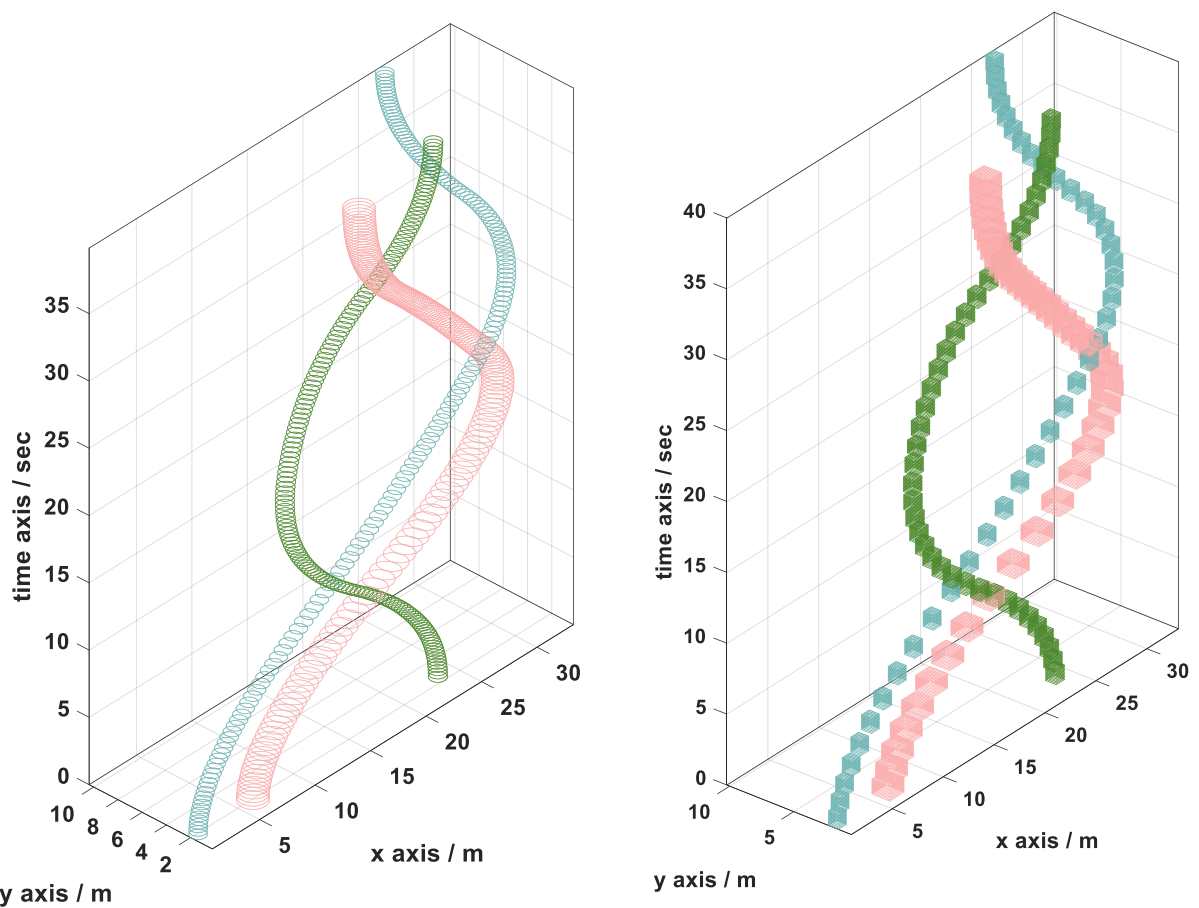

Fig. 3. State space presentations: (a) continuous *x*-*y*-*t* space, and (b) abstracted *x*-*y*-*t* space grids. Note that the colored parts are occupied by moving obstacles.

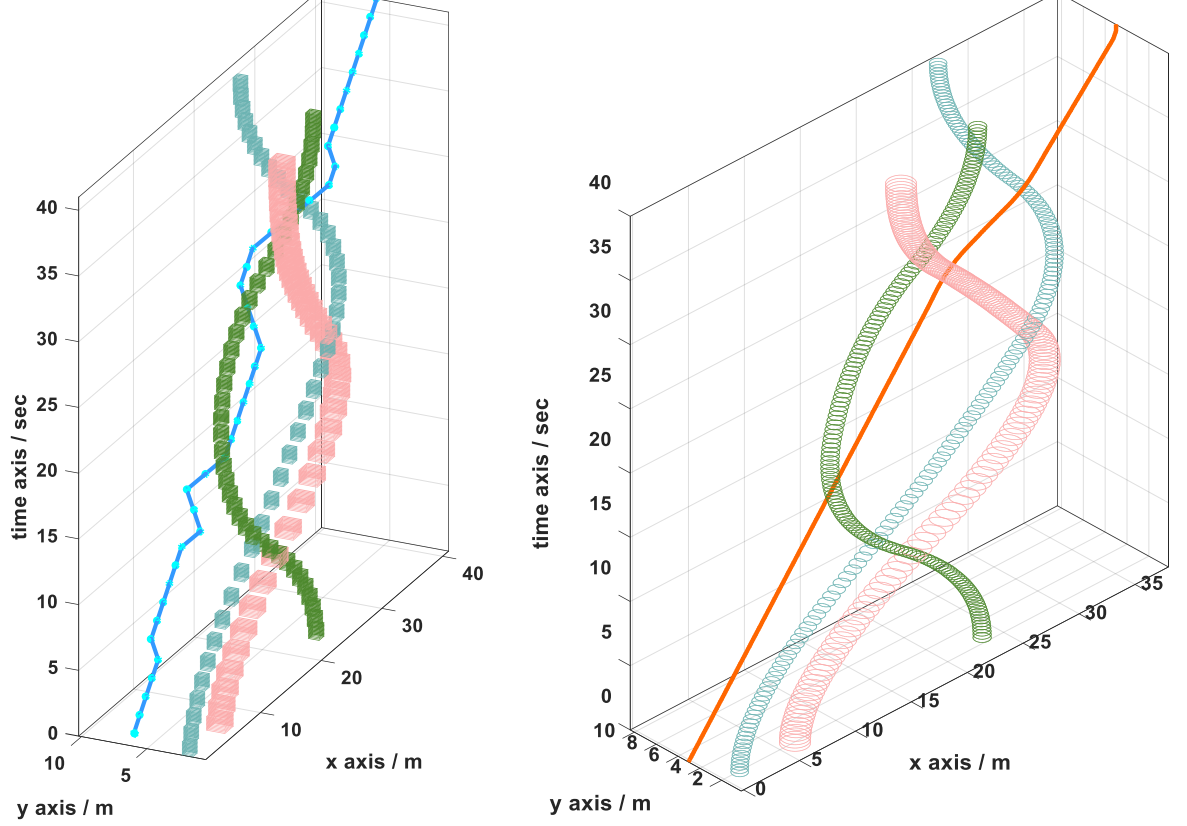

Fig. 4. Trajectory planning results at (a) Stage 1, and (b) Stage 2.

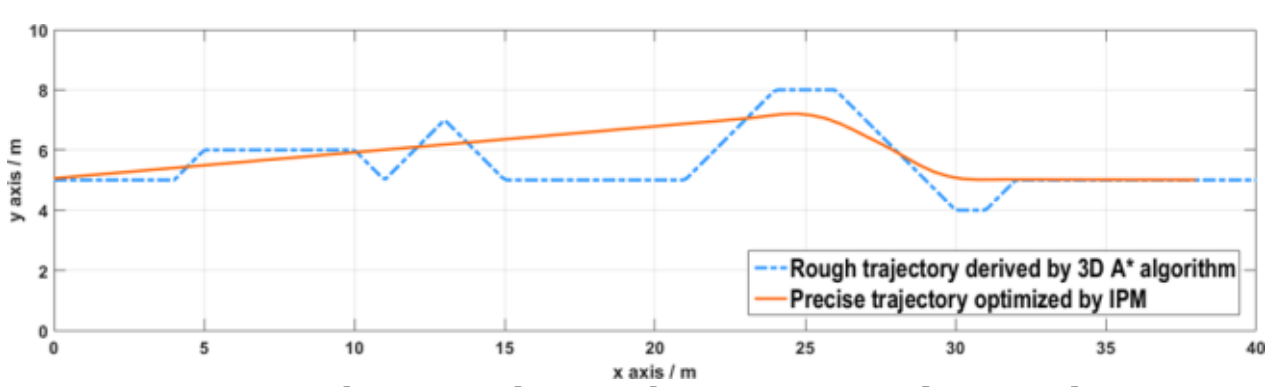

Fig. 5. Comparison between the rough trajectory and precisely optimized trajectory of the ego-AGV.

### B. *On the Time-Consumption of FSTO*

This subsection investigates the computational efficiency of the proposed FSTO. Following the basic setup in Section V.A, we focus on 10 types of scenarios, wherein the number of moving obstacles ranges from 1 to 10. For each type of scenario, 5,000 repeated cases have been tested. In each case, the starting and ending points of the moving obstacles are randomly determined. The results are summarized in Table II.

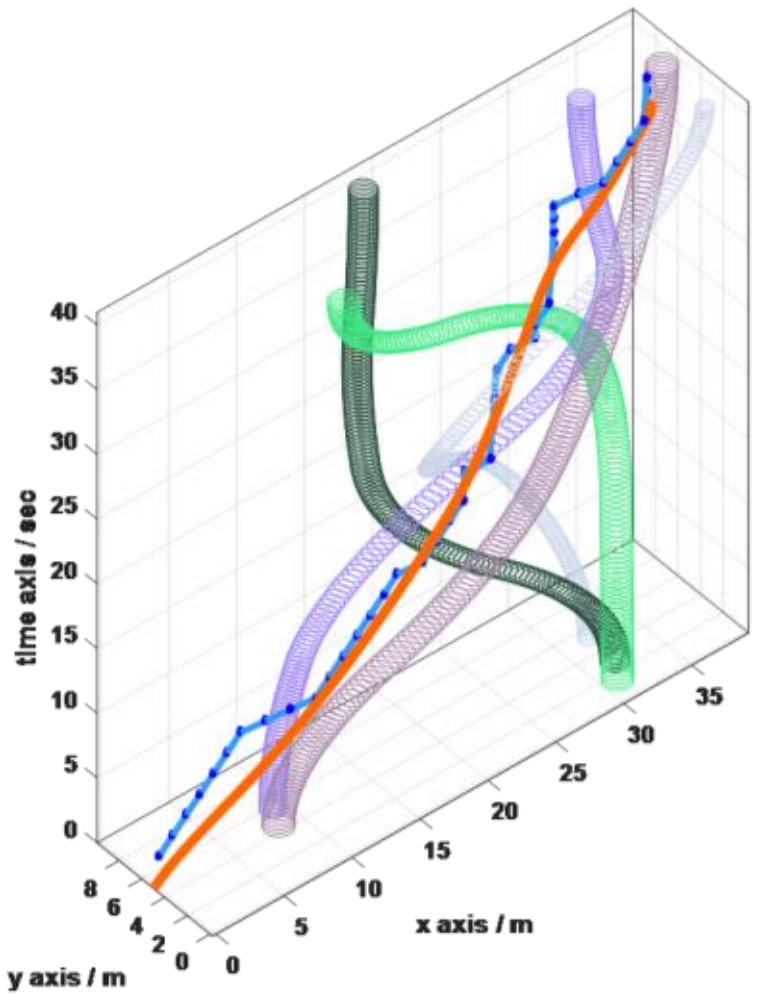

Fig. 6. Illustrations of state space with five moving obstacles, rough trajectory and precisely optimized trajectory.

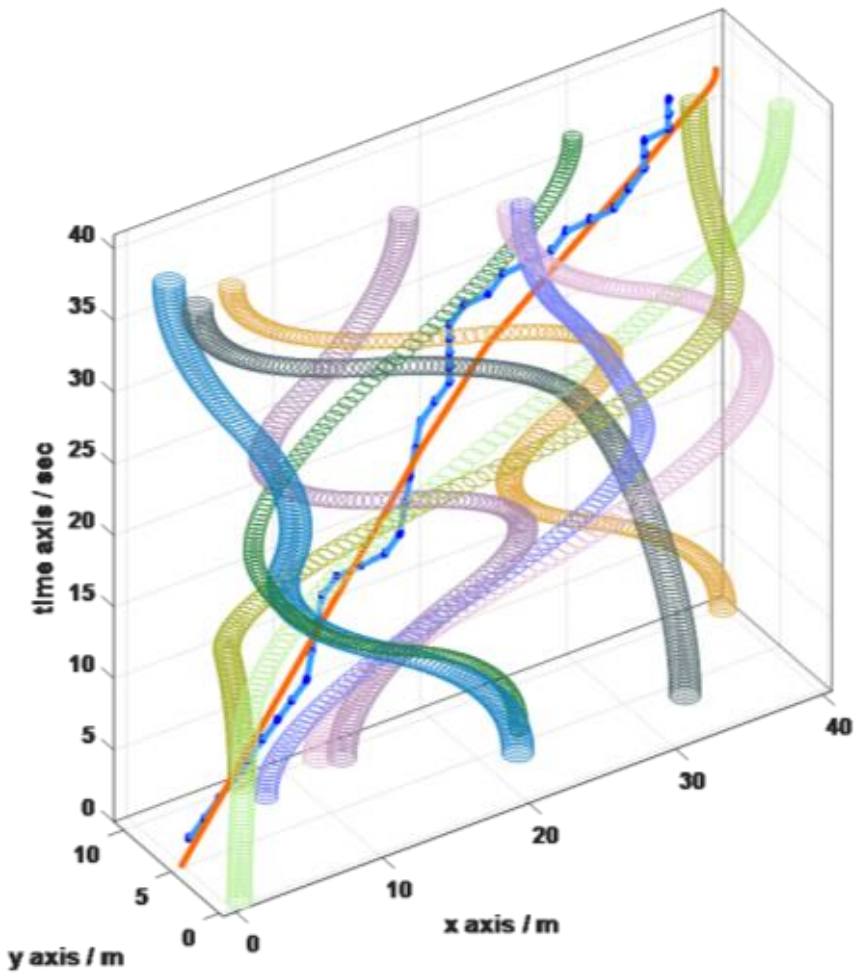

Fig. 7. Illustrations of state space with nine moving obstacles, rough trajectory and precisely optimized trajectory.

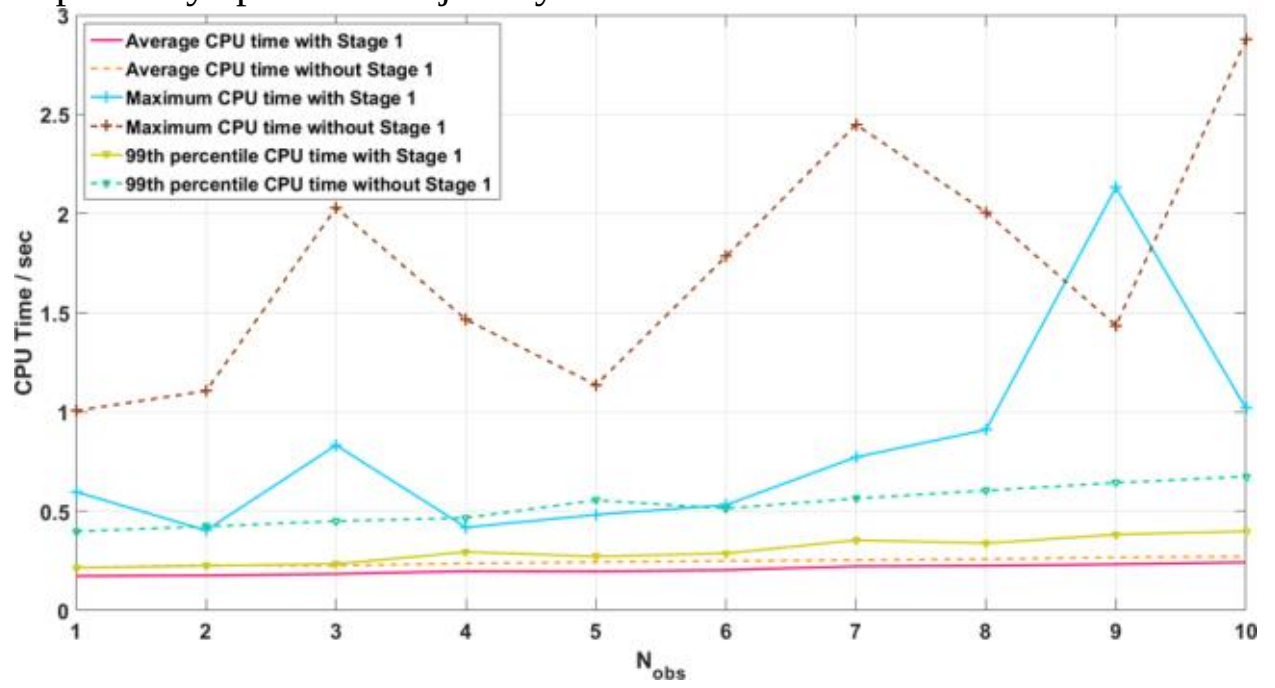

Fig. 8. Comparisons on efficacy of 3D A* algorithm with different $N_{obs}$.

According to Table II, the success rate of 3D A* algorithm gradually decreases as $N_{obs}$ increases. This might due to the fact that the randomly generated case does not have any feasible solution. Among the cases that 3D A* algorithm succeeds, the NLP-solving process at Stage 2 usually completes with high rates of success. Besides that, in either Stage 1 or 2, the CPU time does not significantly grow with $N_{obs}$. Particularly at Stage 1, the scale of the abstracted search space does not change with $N_{obs}$. Through summing up the CPU time in the both stages, one can find that 500 ms is sufficient for most of the cases, which makes the proposed FSTO algorithm capable of handling online trajectory planning schemes.

Recall that the rough trajectory provided at Stage 1 serves as the initial guess, which intends to accelerate the NLP-solving process at Stage 2. In order to investigate the impact of Stage 1 on Stage 2, we have conducted comparative simulations. A competitor of the FSTO approach is defined such that the initial guess is simply generated by connecting the starting and ending points, instead of using the 3D A* algorithm. The simulation results are reported in Table II and further depicted in Fig. 8. FSTO significantly outperforms its competitor regarding all kinds of performance indexes, which indicates that the developed 3D A* algorithm at Stage 1 makes sense to facilitate the NLP-solving process at Stage 2.

### C. *What Can Go Wrong?*

In spite of the efficiency, the proposed FSTO method may typically provide misleading solutions in specific cases. Concretely, although FSTO finds an optimal trajectory that avoids collisions with all of the moving obstacles during $t \in [0, t_f]$, yet it is possible that the ego-AGV stays too closely to the obstacles at $t = t_f$ such that risks are inevitable when $t > t_f$. Frequent replanning with long time horizon helps to address this issue.

## VI. Conclusions

This paper has introduced a unified trajectory planning methodology to deal with moving obstacles with known trajectories. Compared with the prevalent methods, the proposal algorithm can handle a wider range of cases, wherein the trajectories of the moving obstacles may not be straight or piecewise constant. This is achieved through formulating a unified and accurate optimal control problem for case description. In order to facilitate the numerical solving process of the formulated optimal control problem, a graph-search based initialization approach, i.e. the 3D A* algorithm, has been developed. Exhausted simulation results indicate that the proposed first-search-then-optimization (FSTO) strategy is efficient and runs fast.

Regarding our future works, sensor range limits, exterior/interior uncertainties deserve to be considered. Real experiments will be conducted as well.

## Appendix

Source codes of this work are provided at https://github.com/libai1943/AGV_Motion_Planning_with_Moving_Obstacles.

TABLE II. TIME CONSUMPTIONS IN FSTO.

| $N_{obs}$ | **Stage 1** | | | | **Stage 2** | | | | **Simple Initialization without Stage 1** | | | |
|---|---|---|---|---|---|---|---|---|---|---|---|---|
| | Success rate | Average CPU sec | Maximum CPU sec | 99th percentile CPU sec | Success rate | Average CPU sec | Maximum CPU sec | 99th percentile CPU sec | Success rate | Average CPU sec | Maximum CPU sec | 99th percentile CPU sec |
| 1 | 99.720% | 0.033 | 0.269 | 0.041 | 100.000% | 0.140 | 0.328 | 0.173 | 99.619% | 0.215 | 1.006 | 0.398 |
| 2 | 99.620% | 0.033 | 0.109 | 0.042 | 100.000% | 0.143 | 0.294 | 0.183 | 98.510% | 0.228 | 1.106 | 0.423 |
| 3 | 99.280% | 0.035 | 0.053 | 0.043 | 99.980% | 0.148 | 0.779 | 0.191 | 97.138% | 0.226 | 2.029 | 0.450 |
| 4 | 98.960% | 0.038 | 0.084 | 0.051 | 100.000% | 0.160 | 0.334 | 0.243 | 95.154% | 0.236 | 1.465 | 0.466 |
| 5 | 98.300% | 0.039 | 0.074 | 0.049 | 100.000% | 0.158 | 0.408 | 0.223 | 92.231% | 0.243 | 1.136 | 0.555 |
| 6 | 97.980% | 0.040 | 0.077 | 0.050 | 100.000% | 0.163 | 0.454 | 0.237 | 89.465% | 0.249 | 1.784 | 0.513 |
| 7 | 97.680% | 0.043 | 0.105 | 0.067 | 99.959% | 0.179 | 0.667 | 0.287 | 85.075% | 0.254 | 2.448 | 0.564 |
| 8 | 97.140% | 0.045 | 0.090 | 0.060 | 100.000% | 0.180 | 0.820 | 0.278 | 81.967% | 0.258 | 2.006 | 0.604 |
| 9 | 96.980% | 0.046 | 0.159 | 0.061 | 100.000% | 0.186 | 1.973 | 0.321 | 79.210% | 0.267 | 1.436 | 0.643 |
| 10 | 96.840% | 0.048 | 0.100 | 0.069 | 99.979% | 0.194 | 0.918 | 0.329 | 75.067% | 0.271 | 2.877 | 0.674 |